\documentclass[letterpaper, 10 pt, conference]{ieeeconf}  

\IEEEoverridecommandlockouts                              

\usepackage{algorithmic}
\usepackage{rotating}
\usepackage{graphicx}
\usepackage{textcomp}
\usepackage[ruled, algo2e]{algorithm2e}
\usepackage{xcolor}
\usepackage{amsmath,amssymb,amsfonts}

\title{\LARGE \bf
Multi-objective Binary Coordinate Search for Feature Selection
}

\author{Sevil Zanjani Miyandoab$^{*1}$, Shahryar Rahnamayan$^{*2}$, SMIEEE, Azam Asilian Bidgoli$^{*3}$
\thanks{*Nature-Inspired Computational Intelligence (NICI) Lab}
\thanks{$^1$Department of Electrical, Computer, and Software Engineering, Ontario Tech University, Oshawa, 
ON, Canada
        {\tt\small sevil.zanjanimiyandoab@ontariotechu.net}}%
\thanks{$^2$Department of Engineering, Brock University, St. Catharines, ON, Canada
        {\tt\small srahnamayan@brocku.ca}}%
\thanks{$^3$Faculty of Science, Wilfrid Laurier University, Waterloo, ON, Canada
        {\tt\small Abidgoli@wlu.ca}}%
}

\graphicspath{ {./images/} }

\usepackage{tikz}
\usepackage{textcomp}
\usepackage{hyperref}
\usepackage{lipsum}

\newcommand\copyrighttext{%
  \footnotesize \textcopyright 2023 IEEE. Personal use of this material is permitted.
  Permission from IEEE must be obtained for all other uses, in any current or future
  media, including reprinting/republishing this material for advertising or promotional
  purposes, creating new collective works, for resale or redistribution to servers or
  lists, or reuse of any copyrighted component of this work in other works.
  DOI: \href{https://ieeexplore.ieee.org/abstract/document/10394067}{10.1109/SMC53992.2023.10394067}}
\newcommand\copyrightnotice{%
\begin{tikzpicture}[remember picture,overlay]
\node[anchor=south,yshift=10pt] at (current page.south) {\fbox{\parbox{\dimexpr\textwidth-\fboxsep-\fboxrule\relax}{\copyrighttext}}};
\end{tikzpicture}%
}

\begin{document}

\maketitle
\copyrightnotice
\thispagestyle{empty}
\pagestyle{empty}

\begin{abstract}

A supervised feature selection method selects an appropriate but concise set of features to differentiate classes, which is highly expensive for large-scale datasets. Therefore, feature selection should aim at both minimizing the number of selected features and maximizing the accuracy of classification, or any other task. However, this crucial task is computationally highly demanding on many real-world datasets and requires a very efficient algorithm to reach a set of optimal features with a limited number of fitness evaluations. For this purpose, we have proposed the binary multi-objective coordinate search (MOCS) algorithm to solve large-scale feature selection problems. To the best of our knowledge, the proposed algorithm in this paper is the first multi-objective coordinate search algorithm. In this method, we generate new individuals by flipping a variable of the candidate solutions on the Pareto front. This enables us to investigate the effectiveness of each feature in the corresponding subset. In fact, this strategy can play the role of crossover and mutation operators to generate distinct subsets of features. The reported results indicate the significant superiority of our method over NSGA-II, on five real-world large-scale datasets, particularly when the computing budget is limited. Moreover, this simple hyper-parameter-free algorithm can solve feature selection much faster and more efficiently than NSGA-II. 

\end{abstract}

\section{INTRODUCTION}

Feature selection involves eliminating as many features or variables as possible without decreasing the accuracy of classification, or any other task on a specific dataset. Removing irrelevant or redundant features not only reduces computational costs but also improves the classifier's performance by reducing the curse of dimensionality, and also, makes the model easier to interpret \cite{chandrashekar2014survey,ghalwash2016structured}. 

Since over the past years, the domain of features used in machine learning and pattern recognition has broadened to include thousands and millions of variables, the importance of feature selection methods has become more prominent \cite{chandrashekar2014survey}. Each function call in a feature selection task consists of a classification task using a classifier, e.g., k-Nearest Neighbor (kNN), Support Vector Machine (SVM), or Decision Tree (DT), etc. This makes the problem extremely expensive when the dimensionality increases. In this condition, reaching an acceptable result earlier, in a limited computational time or budget, would be greatly appreciated. Besides, reducing the amount of data by reducing the number of features in such large-scale datasets will be desirable. However, eliminating too many features leads to a failure in classification. Hence, we have two conflicting objectives in multi-objective feature selection, namely, the number of features and the accuracy of classification\cite{bidgoli2020evolutionary}.

Feature selection is usually considered as a critical problem with broad applications across different fields, such as bioinformatics (e.g., for nucleotide or amino acid sequence analysis or micro-array analysis) \cite{saeys2007review}, text mining (e.g., for text categorization) \cite{aghdam2009text}, and image analysis (e.g., to find the most appropriate pixels, color, etc) \cite{bidgoli2022evolutionary,asilian2022bias} and many other branches and subjects.

 Feature selection techniques can be categorized into three main groups: wrapper, filter, and embedded methods \cite{agrawal2021metaheuristic,saeys2007review,guyon2003introduction,chandrashekar2014survey,liu2010feature}. The filter-based method is based on analyzing the general characteristics of data and evaluating features without involving any learning models \cite{liu2010feature}. Embedded methods involve variable selection as part of the training process, varying for each learning machine. Using wrappers, subsets of variables are scored based on their predictive power under the black-box of the learning machine \cite{guyon2003introduction,agrawal2021metaheuristic}. 

 Coordinate Search (CS), and similarly, Coordinate Descent (CD), is one of the simplest iterative methods, both in specifying the search direction and updating the variables in each iteration, that solves optimization problems. 
 Reference \cite{ghalwash2016structured} has formulated feature selection as a binary-constrained optimization problem, and proves that coordinate gradient descent is a simple technique that is surprisingly efficient and scalable in variable elimination. Park et al. \cite{park2015modified} have proposed a modified Coordinate Descent methodology by applying changes to its search initialization and adding coordinate randomization as an exploratory step and box search for fine-tuning. In \cite{ghalwash2016structured}, Ghalwash et al. have introduced a fast block coordinate gradient descent method for high dimensional feature selection of structured features grouped according to prior knowledge. Reference \cite{farsa2021population} shows that exploiting a population of individuals rather than a single one further improves the results of this algorithm. As Wright \cite{wright2015coordinate} points out, less attention is sometimes paid to CD algorithms due to their simplicity and lack of sophistication. However, several studies demonstrate their impressive performance \cite{rahnamayan2020towards,feraidooni2020coordinate}. Despite the simplicity and efficiency of coordinate search, there is no multi-objective version of this algorithm to solve an optimization problem with two or more conflicting objectives.

In this study, we propose a binary Multi-Objective Coordinate Search (MOCS) algorithm for the first time, apply it to feature selection as a multi-objective optimization problem, and compare the performance of this approach with the well-known multi-objective evolutionary algorithm, NSGA-II (Non-dominated Sorting Genetic Algorithm II) \cite{deb2002fast}.

To convert feature selection into an optimization problem, the status of each feature would be represented by a binary variable. Since each feature of the dataset can only be kept or removed, each variable has only two states. As a result, each individual - or solution - represents a distinct state of features. The accuracy of classification, which is calculated by masking the features of the dataset and using a regular classifier, e.g., kNN, SVM, or DT, can be considered as an objective value of each individual. Additionally, due to the importance of the number of remaining features for large-scale databases, we define the ratio of the retained features as the second objective of each individual in our optimization problem.

Reference\cite{jiao2022solving} has introduced a multi-objective optimization method for feature selection in classification, called PRDH. They have developed a duplication handling method to enhance the diversity of the population in the objective and search spaces. In addition, their method reformulates the multi-objective feature selection problem as a constrained optimization problem by focusing on classification performance. And a novel constraint-handling method has been used to select feature subsets with more informative and strongly relevant features. Cheng et al. \cite{cheng2022variable} have shown that the novel variable granularity search-based multi-objective evolutionary algorithm, named VGS-MOEA, significantly reduces the search space of large-scale feature selection problems. In this method, each bit of the individuals represents a feature subset. This subset (search granularity) is larger at the beginning and becomes refined gradually, aiming for higher-quality features. This technique can remarkably improve feature selection efficiency and accuracy.

Wang et al. \cite{wang2022information} 
have proposed information theory-based non-dominated sorting ant-colony optimization algorithm, INSA, for bi-objective feature selection. They have introduced a heuristic information function, which improves search efficiency and addresses the imbalance preference problem for the objectives. Moreover, their novel technique of pheromone updating enhances the balance between diversity and convergence.
In order to enhance the evaluation and selection of solutions for a multi-objective optimization problem, a novel fitness evaluation mechanism (FEM) is introduced in \cite{he2021multiobjective}, employing fuzzy relative entropy (FRE). A multi-objective optimization framework is also devised, integrating the FEM with an adaptive local search strategy. To solve the problem effectively, a hybrid genetic algorithm is employed within this framework.

To the best of our knowledge, our proposed method is the first time CS has been modified to apply to multi-objective problems, e.g., feature selection. This is simple in terms of concept and implementation. Although it is a population-based version of CS, it generates distinct sets of solutions in each iteration by changing the status of a feature. A non-dominated sorting algorithm will preserve the most suitable set of features. With this simple scheme of generating operator, it lacks hyper-parameters and search components (such as mutation and crossover), which is a big advantage over many existing multi-objective algorithms. Because of the exponentially reducing search space, convergence can happen in early iterations. It makes this method an excellent candidate for expensive optimization problems with a budget limitation. Moreover, it can be hybridized with other meta-heuristic algorithms, to further improve their performance.

In order to provide a comprehensive understanding of the topic and to present our research methodology and findings in a logical manner, this paper is organized as follows: Section II provides a detailed background review, Section III outlines our proposed method, highlighting the steps and employed techniques, Section IV showcases results and analysis derived from our study, and Section V concludes with remarks.

\section{BACKGROUND REVIEW}
Two main building components of the proposed method are Coordinate Search and Multi-objective Optimization which are explained in detail in the following subsections.  
\subsection{Coordinate Search}
Coordinate descent (CD) algorithms try to tackle optimization problems by solving a sequence of simple optimization problems~\cite{schwefel1993evolution}. Thus, they are a category of decomposition-based algorithms. At each iteration, CD optimizes one or a block of coordinates (variables) while fixing all other coordinates or blocks~\cite{bidgoli2021memetic}. In other words, the idea behind the CD algorithm is that, when directional derivatives are unavailable or difficult to compute, one-dimensional minimization can be replaced to approximate a good solution~\cite{frandi2014coordinate}. In numerical linear algebra or arithmetic optimization, gradient information is required to apply CD. However, in the Evolutionary Computation community, when exact gradient information is not applicable, the method is called coordinate search (CS). In this case, the algorithm benefits function value samplings on coordinates in turn to find a suitable value for each variable. 

Despite its simplicity and inexact derivative-free minimization, the algorithm still produces acceptable practical results and performs better than other algorithms in solving expensive optimization problems such as feature selection. Depending on several parameters which determine the framework of the CS algorithm, different versions of the algorithm are proposed. Some of these parameters are the order of coordinates or blocks to be optimized in turn, the number of simultaneously updated coordinates, the initial point to start the optimization process, the number of sampled points, and the way that sampling is conducted (i.e., the step value of sampling).  

Depending on the number of coordinates that are optimized simultaneously, CS can be viewed as two variants. At each iteration, the algorithm may update only one coordinate based on the fitness evaluation of the sampled points while all other coordinates are fixed. In order to accelerate the optimization process, particularly in large- or huge- scale optimization problems, a block-CS can be utilized in which instead of only one coordinate, a block of variables can be updated simultaneously, and consequently, the number of fitness evaluations is decreased dramatically~\cite{tseng2001convergence}.

\subsection{Multi-objective Optimization}
Multi-objective optimization has been defined as optimizing two or more conflicting objectives.The algorithms to solve these problems usually make a trade-off decision and generate a set of solutions instead of only one. This set of solutions is called the Pareto front, or non-dominated solutions. The Pareto front is created using the concept of dominance, which is used to compare solutions. 

\textbf{Definition 1. Multi-objective Optimization}~\cite{asilian2022machine}
\begin{eqnarray}
\begin{aligned}
& Min/Max\  F(\pmb x)=[f_{1}(\pmb x),f_{2}(\pmb x),...,f_{M}(\pmb x)] \\ 
&s.t. \quad L_{i}\leq x_{i}\leq U_{i}, i=1,2,...,d
 \end{aligned}
\end{eqnarray}
where $M$ is the number of objectives, $d$ is the number of decision variables (i.e., dimension), and the value of each variable, $\pmb x_{i}$, is in the interval $[L_{i}, U_{i}]$ (i.e., box-constraints). $f_{i}$ represents the objective function, which should be minimized or maximized.

One of the commonly used  concepts for comparing candidate solutions in such problems is dominance.

\textbf{Definition 2. Dominance Concept} \\
If  $\pmb x=(x_{1},x_{2},...,x_{d})$ and  $ \acute{\pmb x}=(\acute{x}_{1},\acute{x}_{2},...,\acute{ x}_{d})$ are two vectors in a minimization problem search space, $\pmb x$ dominates $\acute{\pmb x}$ ($\pmb x\prec\acute{\pmb x}$) if and only if
\begin{eqnarray}
\begin{aligned}
&\forall i\in{\{1,2,...,M\}}, f_i(\pmb x)\leq f_i(\acute{\pmb x}) \wedge\\ 
&\exists j \in{\{1,2,...,M\}}: f_j(\pmb x)<f_j(\acute{\pmb x})
\end{aligned}
\end{eqnarray}
This concept defines the optimality of a solution in a multi-objective space. Candidate solution $\pmb x$ is better than $\acute{\pmb x}$ if it is not worse than $\acute{\pmb x}$ in any of the objectives and at least it has a better value in one of the objectives. All of the solutions, which are not dominated by any other solution, create the Pareto front and are called non-dominated solutions \cite{bidgoli2021reference}.

Another popular method for comparing candidate solutions in multi-objective problems is crowding distance. For computing the crowding distance of the individuals, we sort them based on each objective and assign an infinite distance to the boundary individuals (minimums and maximums). Other individuals are assigned the sum of the normalized Euclidean individual distance from their neighbors with the same rank on all objectives. This estimates the importance of an individual in relation to the density of individuals surrounding it~\cite{deb2002fast}.

 Multi-objective algorithms attempt to find the Pareto front by utilizing generating strategies/operators and selection schemes. The non-dominated sorting (NDS) algorithm~\cite{deb2002fast} is one of the popular selection strategies which works based on the dominance concept. It ranks the solutions of the population in different levels of optimality. The algorithm starts with determining all non-dominated solutions in the first rank.
 
 In order to identify the second rank of individuals, the non-dominated vectors are removed from the set to process the remaining candidate solutions in the same way.
 This process will continue until all of the individuals are grouped into different levels of Pareto. Deb et al. \cite{deb2002fast} has introduced one of the most famous and impressive state-of-the-art multi-objective optimization methods, NSGA-II, based on this concept. It is the main competitor to our method.


\section{PROPOSED METHOD}

\subsection{Objectives}
As mentioned in the previous sections, improving classification accuracy as well as reducing the number of features are the main goals of a large-scale feature selection problem. Therefore, we define two objectives for each candidate solution (or set of features). Our first objective is to increase the accuracy of classification. For calculating the classification accuracy, we employ the k-Nearest Neighbor (kNN) classifier to be trained and evaluated on the train set (during the optimization phase) and test set (after optimization for evaluating the final solutions).

In order to design a minimization optimization problem, we define the error of classification as the first objective using the predictions of the kNN:

\begin{equation}
\label{f1}
\mathit{Classification \:Error}= 1 - \frac{\#  Correct\: Predictions}{Total \:\# Predictions}
\end{equation}

The second objective is to minimize the number of selected features. For that reason, we count the number of variables with the value of 1 (or True) for each individual, and accordingly, we can compute the ratio of the selected features:

\begin{equation}
\label{f2}
\mathit{Ratio\: of\: Selected\: Features}=\frac{Number \:of \: 1's }{Total \:\# Features }
\end{equation}

The optimizer tries to minimize both of these objectives, and both may have a real value in the interval \([0, 1]\). Studies have shown that these two objectives are in conflict, so they are appropriate candidates for multi-objective optimization \cite{bidgoli2020evolutionary}.

\subsection{MOCS}
 In the proposed MOCS for feature selection, each feature is evaluated individually by considering two objectives. For this purpose, at each iteration, we choose a feature based on a random permutation of the variables and change the status of the corresponding feature in all individuals of the population to generate new candidate solutions. This method is basically based on the process of the CS algorithm in which the value of one variable is changed while all others are kept unchanged. 

 Since the goal of the proposed method is to generate a Pareto front for multi-objective feature selection, a population for CS should be created. Although the selection strategy is partly adopted from the NSGA-II algorithm, generating new subsets of features in each generation is a novel approach that can be effectively applied to large-scale feature selection and probably other binary multi-objective tasks.

The proposed method consists of several major steps as mentioned in Algorithm~\ref{alg-one}. The details of each step are provided as follows:

\begin{enumerate}
   \item Initialization: First of all, a random population of $N$ solutions is formed. An individual, also known as a candidate solution or chromosome in the adaptation of optimization with the feature selection problem, consists of strings of bits - 0s and 1s. These variables - or bits - of an individual determine whether corresponding features in the original dataset are selected or removed. If a variable is set to 1, the corresponding feature in the dataset will remain; on the other hand, if the variable is set to 0, the feature will be deleted. Accordingly, each individual - or solution - represents a distinct state of the features. After generating the initial population,  the objective values for each individual (set of features) are calculated. At this point, we have evaluated $N$ individuals so far. Therefore, the number of function calls ($NFC$) should be set to $N$.
   \item Selection/Survival: NDS is applied to the population to identify the Pareto front. Since the next steps are required to be applied to the best set of candidate solutions, dominated candidate solutions are no longer needed and can be removed from the population. 
   \item Generation of New Solutions: We create an empty temporary list. New solutions are generated by flipping a specific variable of the solutions in the Pareto front and calculating their objective values. Then the new candidate solution should be compared with its parent. If a newly generated candidate solution is not dominated by its parent, it will be added to the temporary list. This allows the effectiveness of each feature to be assessed one-by-one.
   \item Formation of the New Population: We merge the individuals in the temporary list (mentioned in the previous step) with the population (parents or current Pareto front).
   \item Selection/Survival: After applying NDS to the population to identify the Pareto front and removing the dominated solutions as in step 2, we may also want to check the population size here. If the number of members of the Pareto front exceeds a maximum threshold, e.g., $N$, we use the crowding distance to select the $N$ best solutions of the Pareto front and eliminate the other individuals from the population.
   \item If the termination condition is not met, a new iteration starts by jumping to step 3 and assessing the next feature. For instance, if the number of function calls ($NFC$) has not reached the maximum allowed number ($maxNFC$), the algorithm should continue in the loop. All these steps should be iterated over all variables (i.e., features). Since the order of features can play a role in the assessment of each feature, after iterating over all variables, a new random permutation is considered on the order of features to  start a subsequent round of assessment. 
\end{enumerate}

\begin{algorithm2e}
\SetAlgoLined
\SetKwInOut{Input}{input}\SetKwInOut{Output}{output}
 \Input{ $dataset$, $N$, $maxNFC$ }
 \Output{ $Pareto front$ }
 \BlankLine
\tcp{Initialization}
$population$ = random population with size $N$\;
evaluate($population$)\;
$NFC = N$\;
$population = NDS(population).front[0]$\;

\While{true}{
$permutation = $ New Permutation\;
\tcp{Iterating over variables}
\For{$i\leftarrow 1$ \KwTo $NumberofFeatures$}{
$index=permutation[i]$\;
\tcp{Generating new individuals}
$NewCandidateSolutions = $ New List\;
\tcp*[h]{Iterating over population}\; 
\For{$j\leftarrow 1$ \KwTo $len(population)$}
    {
    $NewIndividual = population[j]$\;
    $NewIndividual[index] = !NewIndividual[index]$\;
    evaluate($NewIndividual$)\;
    $NFC = NFC + 1$\;
    \If {$population[j]$ does not dominate $NewIndividual$}
          	{add $NewIndividual$ to $NewCandidateSolutions$}
    }

$population= NewCandidateSolutions \cup population$\;
$population = NDS(population).front[0]$\;

\If {$NFC > maxNFC$ or population has not changed in \(2\times D\) iterations}
          	{Exit}

\If{$len(population)>N$} {
calculate the crowding distance of individuals\;
sort individuals based on crowding distance\;
$population = N$ best individuals\;}

}
}
\BlankLine
\caption{Pseudo-code of binary MOCS}\label{alg-one}
\end{algorithm2e}

 In contrast to NSGA-II, our method does not employ crossover and mutation operators. Instead, we alter the state of only one variable of selected individuals to generate distinct children - or better put, candidate solutions. When a newly generated solution is not dominated by its corresponding member of the Pareto front, it will be merged with the population. After this, the survival or selection process would be run on the population and the non-dominated set of solutions would survive. Furthermore, the crowding distance of the individuals is not necessary to be calculated in each iteration. If and only if the Pareto front size exceeds a specific number $N$, we calculate and sort the crowding distances of the solutions and select the best $N$ individuals from the Pareto front.  We set $N$ equal to the initial population size so that the number of individuals never rises beyond $2\times N$ since the number of newly generated solutions can be in the range of \([1, N]\). However, some of them may be eliminated before merging with the current population because of the dominance of their parents.

It is worth mentioning that in most cases, eliminating individuals other than those in the Pareto front in each iteration leads to a decrease in memory usage compared to NSGA-II, which keeps \(2\times N\) individuals in each iteration. Although, our proposed method and NSGA-II have similar time and memory complexities. The most computationally demanding operator in each NSGA-II iteration is NDS with $O(MN^2)$ computations\cite{deb2002fast}. MOCS also includes this selection process, while its novel generation technique does not exceed $O(N)$  computations to generate new candidate solutions. As a result, its computational complexity for each iteration is  $O(MN^2)$.

We may define a termination condition based on the total number of evaluations (as in our experiments), so that, after evaluating a specific number of solutions, the algorithm stops and the output would be the final Pareto front. Another termination option that can be considered for the proposed method is \emph{convergence}. In MOCS, if  the Pareto front remains unchanged  in \(2\times D\) (i.e., two times the total number of features) subsequent iterations, it indicates that the algorithm has converged and it cannot find any better solution. The reason is that after that number of iterations, the status of all features has been flipped and evaluated; therefore, if no improvement has been achieved, the next iterations will not result in distinct solutions. Thus, the algorithm can stop. This is an advantage of the proposed algorithm. While we cannot identify a definite convergence indicator in other stochastic evolutionary methods such as NSGA-II.


\section{EXPERIMENTAL RESULTS AND ANALYSIS}
\subsection{Datasets}
We compare the performance of MOCS and NSGA-II on five large-scale datasets \cite{zhao2010advancing} in the fields of microarray and image/face recognition to select the best set of features. One of the most prominent characteristics of datasets in these fields is their high number of features with a relatively small number of instances. It deteriorates the classifier's performance and increases their need to reduce the number of features. You may find the properties of the adopted datasets in Table~\ref{tab-datasets}.

\subsection{Experimental Settings}
Twenty percent of each dataset's instances are randomly selected as test sets, which are not seen during optimization. Due to the stochasticity of optimization algorithms, we run the algorithm 10 times and at each run, a random subset of samples is considered as the test set. Therefore, the experiments are somehow similar to 10-fold cross-validation with a probability of overlapping the test sets. Because of the expensiveness of the feature selection process on large-scale datasets, we have fixed the number of function calls (maximum NFC), or evaluations, to 50,000 for both algorithms to have a fair comparison. Details of the hyperparameters can be found in Table~\ref{tab-settings}.

We consider the error of classification as the first objective and the ratio of the selected features to all features as the second objective. Consequently, minimization of both objectives is desired and both may have a real value in the interval \([0, 1]\). We evaluate the multi-objective optimization algorithms by measuring the hypervolume (HV) with the reference point of $(1, 1)$.

We use kNN implemented with the help of FAISS \cite{fastKNN} as the classifier on the train set to compute the classification error on each candidate solution (i.e., a subset of features). The $k$ parameter for warpAR10P, warpPIE10P, TOX-171, pixraw10P, and CLL-SUB-111 datasets is 5, 5, 5, 5, and 4, respectively. 

\begin{table}[htbp]
\caption{Datasets Description}
\begin{center}
\begin{tabular}{|c|c|c|c|c|}
\hline
\textbf{Dataset}     & \textbf{\#Features} & \textbf{\#Instances} & \textbf{\#Classes} & \textbf{Domain}          \\ \hline
warpAR10P   & 2400       & 130         & 10        & Image, Face     \\ \hline
warpPIE10P  & 2420       & 210         & 10        & Image, Face     \\ \hline
TOX-171     & 5748       & 171         & 4         & Microarray \\ \hline
pixraw10P   & 10000      & 100         & 10        & Image, Face     \\ \hline
CLL-SUB-111 & 11340      & 111         & 3         & Microarray \\ \hline
\end{tabular}
\label{tab-datasets}
\end{center}
\end{table}

\subsection{Numerical Results and Analysis}

\begin{figure*}
\centering
\begin{tabular}{cc}
\includegraphics[width=0.32\linewidth]{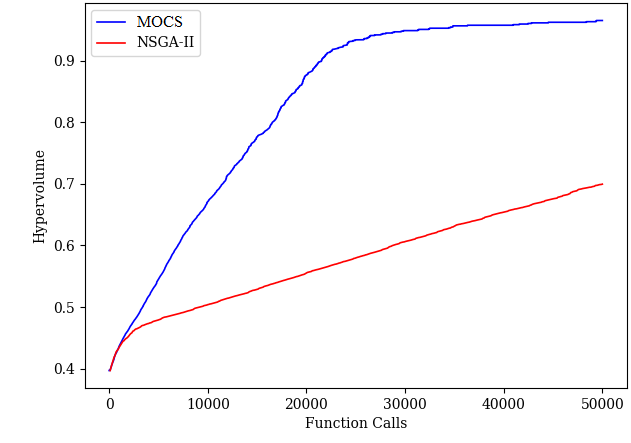} &\includegraphics[width=0.33\linewidth]{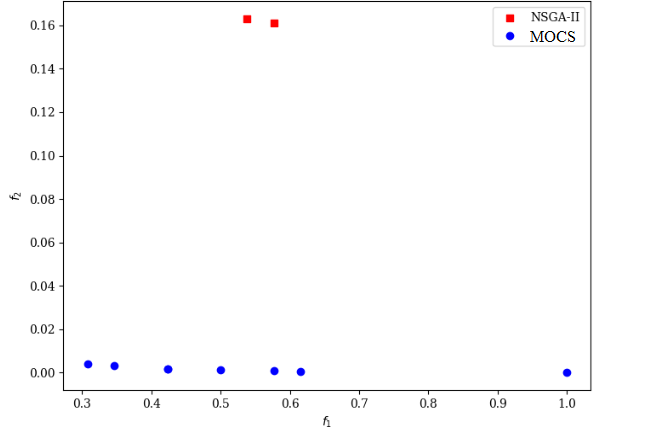}\\
(a) warpAR10P & (b) warpAR10P\\
\includegraphics[width=0.32\linewidth]{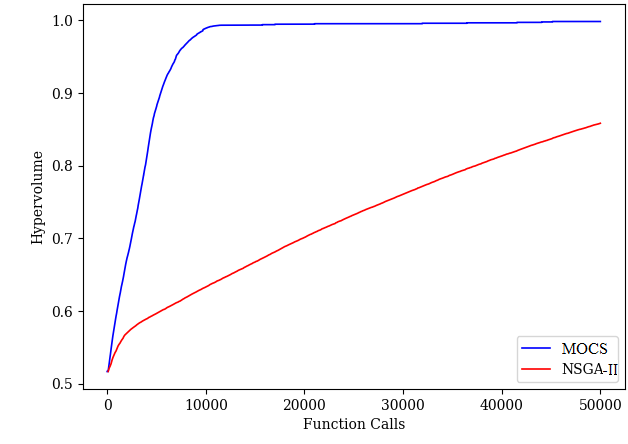} &\includegraphics[width=0.33\linewidth]{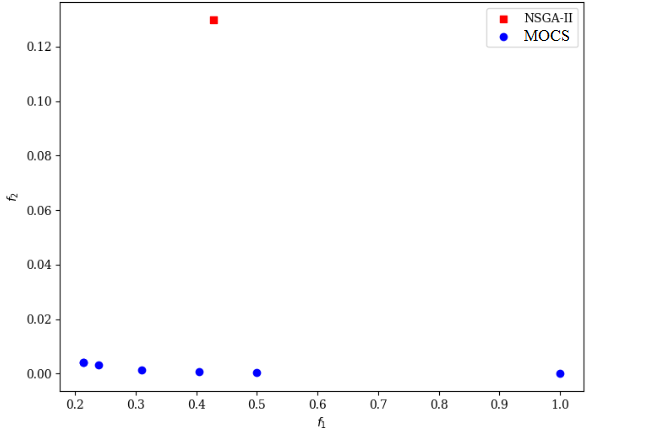}\\
(c) warpPIE10P & (d) warpPIE10P\\
\includegraphics[width=0.32\linewidth]{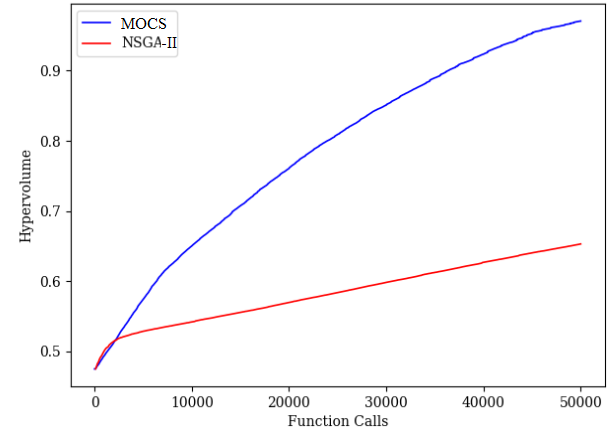} &\includegraphics[width=0.34\linewidth]{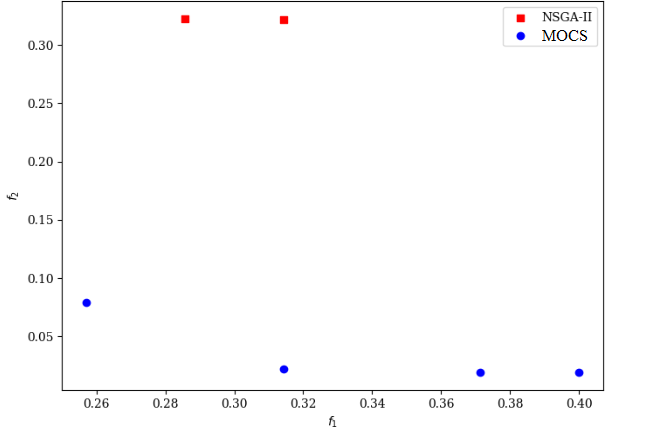}\\
(e) TOX-171 & (f) TOX-171\\
\includegraphics[width=0.32\linewidth]{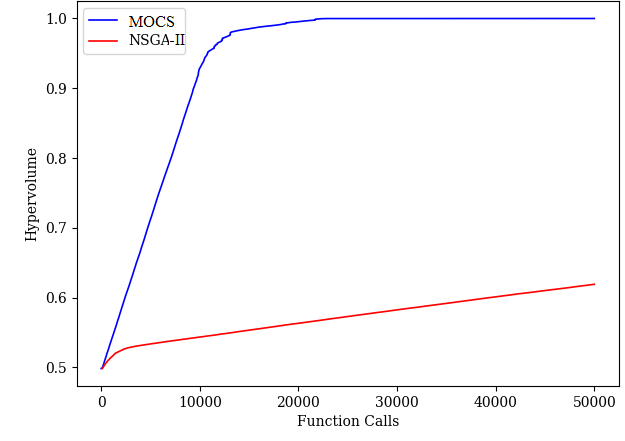} &\includegraphics[width=0.33\linewidth]{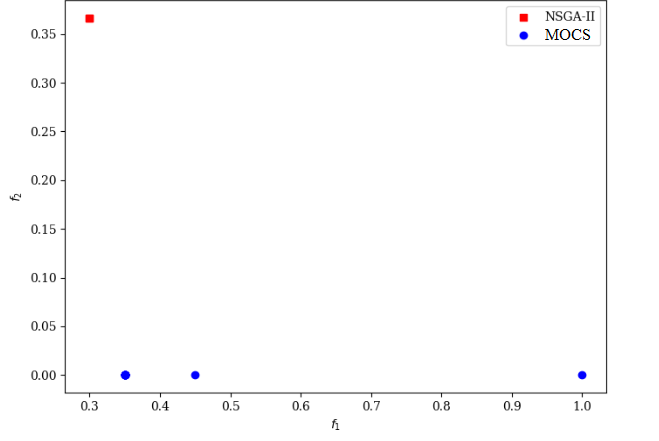}\\
(g) pixraw10P & (h) pixraw10P\\
\includegraphics[width=0.32\linewidth]{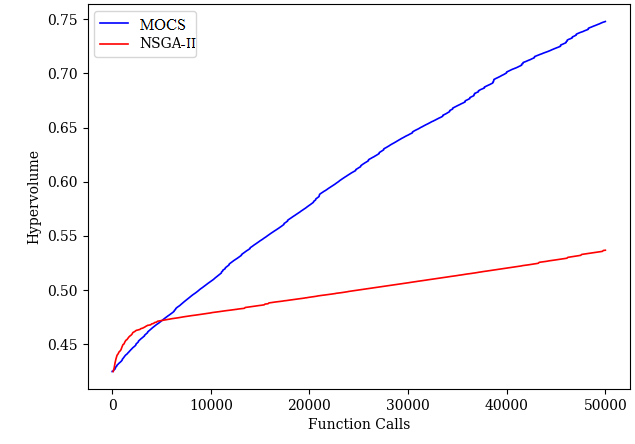} &\includegraphics[width=0.33\linewidth]{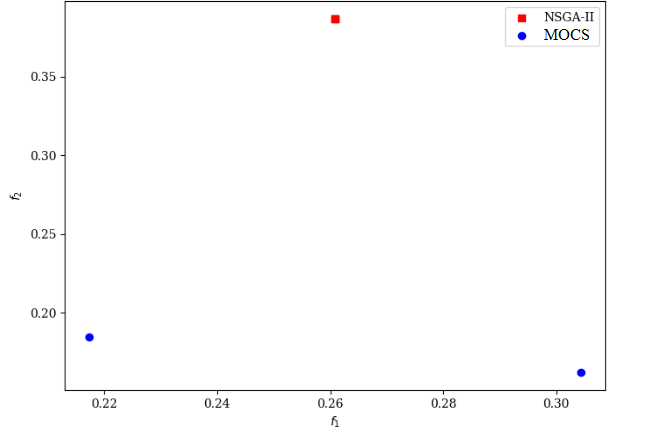}\\
(i) CLL-SUB-111 & (j) CLL-SUB-111\\

\end{tabular}
\caption{HV plots during optimization on the train set (left) and median final test Pareto set (right) on datasets. $f_1$ and $f_2$ are the classification error and the ratio of selected features, respectively.}
\label{img-1}
\end{figure*}

Fig.~\ref{img-1} shows the average trend of the HV during optimization for both methods (left column) and the resultant Pareto fronts on test sets (right column). We see that in all datasets, MOCS presents a very rapid jump-up of HV, while the HV plots of NSGA-II reach a relatively low value in such large-scale search spaces and cannot improve significantly compared to MOCS. Evolutionary algorithms have lower improvement rates than CS for solving large-scale optimization problems.

Table~\ref{tab-results-HV} contains the average of the numerical results for the final Pareto fronts resulting from MOCS and NSGA-II algorithms after 50,000 function calls. As you can see in the table, numerical results in every aspect prove the superiority of MOCS over NSGA-II. In all datasets, the final train and test HV resulting from MOCS have a significant advantage over NSGA-II. The HV rises much slower with NSGA-II than with MOCS, as shown in Fig.~\ref{img-1}. This is especially beneficial and valuable for computationally expensive optimization problems as with this limited number of fitness calls, MOCS can quickly reach a set of optimal features. MOCS reaches a train HV of 0.97 for 4 datasets out of 5, while NSGA-II cannot reach a train HV of 0.87 for any datasets. For datasets in the field of image or face recognition, both algorithms increase HV faster than microarray in most cases. Optimizing a large microarray dataset such as CLL-SUB-111 requires more function calls even for MOCS. In order to demonstrate the improvement of HV values during the optimization process, the initial values of HV which are computed from the initial population are also reported. Not only is the HV of the train Pareto front of MOCS significantly higher than NSGA-II, but also the Pareto front of test data results in superior HV values than NSGA-II. 

Moreover, Table~\ref{tab-results-Obj} represents the minimum classification error and the average ratio of features on the final train Pareto fronts for MOCS and NSGA-II. In addition, the resultant Pareto fronts on test sets demonstrate better sets of features in terms of both objectives. In four datasets, the average number of features of the Pareto front has been reduced to 2\% of the total features or less. While NSGA-II keeps more than 12\% of the features for all of the datasets after a similar number of function calls. Therefore, we notice that MOCS is exceptionally successful at eliminating unnecessary features and reducing the ratio of retained features (the second objective) while minimizing classification error (the first objective). 

MOCS not only remarkably improves the simultaneous minimization of both objectives, but also makes the distribution of Pareto front solutions in the objective space (mostly the first objective: classification error) broader in most cases. A wider distribution of solutions provides decision-makers with more options in practice. MOCS' solutions are rarely worse than NSGA-II's solutions. 

An earlier convergence of MOCS increases the number of solutions on the Pareto front. For instance, after 20,000 function calls for the warpPIE10P dataset, the HV almost does not change in MOCS, while exploring the features of the population leads to finding new solutions on the Pareto front or even with duplicate objective values (we eliminate solutions with a duplicate set of variables for both algorithms). Consequently, some solutions may overlap in the objective space. For this reason, the number of MOCS Pareto front solutions for this dataset is remarkably more than NSGA-II's; on the other hand, this difference is much less for the CLL-SUB-111 dataset, in which both algorithms have not converged yet. 

Consequently, the proposed binary MOCS can efficiently solve huge- and large-scale optimization problems such as feature selection, which are computationally demanding. Allocating numerous fitness evaluations to regular population-based evolutionary algorithms is not usually applicable to real-world problems. Another advantage of MOCS is that it is a hyper-parameter-free algorithm. Dislike many evolutionary algorithms which can be highly affected by tuning the hyper-parameters, the only parameter that should be considered in MOCS is the initial population size. However, even this parameter loses its impact during the process.

\begin{table}[htbp]
\caption{Parameter Settings for both MOCS and NSGA-II algorithms}
\begin{center}
\begin{tabular}{|ll|}
\hline
\multicolumn{2}{|c|}{\textbf{MOCS}}                                                    \\ \hline
\multicolumn{1}{|l|}{Maximum size of population}     & 100                    \\ \hline
\multicolumn{1}{|l|}{Number of function calls (NFC)} & 50000                  \\ \hline
\multicolumn{1}{|l|}{Sampling method}                & Binary Random Sampling \\ \hline
\multicolumn{1}{|l|}{Survival method}                & NDS algorithm          \\ \hline
\multicolumn{1}{|l|}{Duplicate Elimination}          & TRUE                   \\ \hline
\multicolumn{1}{|l|}{Number of runs of algorithm}    & 10                     \\ \hline
\multicolumn{2}{|c|}{\textbf{NSGA-II}}                                                 \\ \hline
\multicolumn{1}{|l|}{Population size}                & 100                    \\ \hline
\multicolumn{1}{|l|}{Number of function calls (NFC)} & 50000                  \\ \hline
\multicolumn{1}{|l|}{Sampling method}                & Binary Random Sampling \\ \hline
\multicolumn{1}{|l|}{Selection method}               & Tournament Selection   \\ \hline
\multicolumn{1}{|l|}{Mutation method}                & Bit-flip Mutation      \\ \hline
\multicolumn{1}{|l|}{Crossover}                      & SPX                    \\ \hline
\multicolumn{1}{|l|}{Survival method}                & NDS algorithm          \\ \hline
\multicolumn{1}{|l|}{Duplicate Elimination}          & TRUE                   \\ \hline
\multicolumn{1}{|l|}{Number of runs of algorithm}    & 10                     \\ \hline
\end{tabular}
\label{tab-settings}
\end{center}
\end{table}

\begin{table*}[htbp]
\caption{Results for comparing average final training and testing HV and number of solutions in the Pareto fronts of MOCS and NSGA-II. Initial HV is computed over the initial population. }
\begin{center}
\begin{tabular}{|c|c|ccc|ccc|}
\hline
                 &                     & \multicolumn{3}{c|}{\textbf{MOCS}}                                                                                                                                                                                                                        & \multicolumn{3}{c|}{\textbf{NSGA-II}}                                                                                                                                                                                                                     \\ \hline
\textbf{Dataset} & \textbf{Initial HV} & \multicolumn{1}{c|}{\textbf{\begin{tabular}[c]{@{}c@{}}Final\\ Train HV\end{tabular}}} & \multicolumn{1}{c|}{\textbf{\begin{tabular}[c]{@{}c@{}}Final\\ Test HV\end{tabular}}} & \textbf{\begin{tabular}[c]{@{}c@{}}\#Solutions\\ in Pareto front\end{tabular}} & \multicolumn{1}{c|}{\textbf{\begin{tabular}[c]{@{}c@{}}Final\\ Train HV\end{tabular}}} & \multicolumn{1}{c|}{\textbf{\begin{tabular}[c]{@{}c@{}}Final\\ Test HV\end{tabular}}} & \textbf{\begin{tabular}[c]{@{}c@{}}\#Solutions\\ in Pareto front\end{tabular}} \\ \hline
warpAR10P        & 0.27±0.05           & \multicolumn{1}{c|}{\textbf{0.97±0.01}}                                                & \multicolumn{1}{c|}{\textbf{0.67±0.05}}                                               & \textbf{20.1±5.2}                                                        & \multicolumn{1}{c|}{0.70±0.03}                                                         & \multicolumn{1}{c|}{0.45±0.09}                                                        & 4.3±2.3                                                                  \\ \hline
warpPIE10P       & 0.35±0.03           & \multicolumn{1}{c|}{\textbf{0.99±0.01}}                                                & \multicolumn{1}{c|}{\textbf{0.78±0.07}}                                               & \textbf{45.4±23.7}                                                       & \multicolumn{1}{c|}{0.86±0.01}                                                         & \multicolumn{1}{c|}{0.57±0.04}                                                        & 2.9±3.0                                                                  \\ \hline
TOX-171          & 0.38±0.04           & \multicolumn{1}{c|}{\textbf{0.97±0.01}}                                                & \multicolumn{1}{c|}{\textbf{0.73±0.07}}                                               & \textbf{24.1±5.5}                                                        & \multicolumn{1}{c|}{0.65±0.02}                                                         & \multicolumn{1}{c|}{0.49±0.04}                                                        & 2.5±1.9                                                                  \\ \hline
pixraw10P        & 0.35±0.07           & \multicolumn{1}{c|}{\textbf{0.99±0.01}}                                                & \multicolumn{1}{c|}{\textbf{0.70±0.11}}                                               & \textbf{59.6±33.9}                                                       & \multicolumn{1}{c|}{0.62±0.01}                                                         & \multicolumn{1}{c|}{0.43±0.07}                                                        & 2.5±2.8                                                                  \\ \hline
CLL-SUB-111      & 0.36±0.05           & \multicolumn{1}{c|}{\textbf{0.75±0.06}}                                                & \multicolumn{1}{c|}{\textbf{0.62±0.09}}                                               & \textbf{11.4±1.4}                                                        & \multicolumn{1}{c|}{0.54±0.01}                                                         & \multicolumn{1}{c|}{0.42±0.05}                                                        & 3.5±2.8                                                                 \\ \hline
\textbf{Average} & 0.34±0.05           & \multicolumn{1}{c|}{\textbf{0.93±0.02}}                                                & \multicolumn{1}{c|}{\textbf{0.70±0.08}}                                               & \textbf{32.1±13.9}                                                       & \multicolumn{1}{c|}{0.67±0.02}                                                         & \multicolumn{1}{c|}{0.47±0.06}                                                        & 3.1±2.6                                                                  \\ \hline
\end{tabular}
\label{tab-results-HV}
\end{center}
\end{table*}

\begin{table*}[htbp]
\caption{Results for comparing average final objective values in the Pareto fronts of MOCS and NSGA-II}
\begin{center}\
\begin{tabular}{|c|cc|cc|}
\hline
                 & \multicolumn{2}{c|}{\textbf{MOCS}}                                                                                                                                           & \multicolumn{2}{c|}{\textbf{NSGA-II}}                                                                                                                                        \\ \hline
\textbf{Dataset} & \multicolumn{1}{c|}{\textbf{\begin{tabular}[c]{@{}c@{}}Minimum\\ Classification Error\end{tabular}}} & \textbf{\begin{tabular}[c]{@{}c@{}}Average\\ Ratio of Features\end{tabular}} & \multicolumn{1}{c|}{\textbf{\begin{tabular}[c]{@{}c@{}}Minimum\\ Classification Error\end{tabular}}} & \textbf{\begin{tabular}[c]{@{}c@{}}Average\\ Ratio of Features\end{tabular}} \\ \hline
warpAR10P        & \multicolumn{1}{c|}{\textbf{0.03±0.01}}                                                              & \textbf{0.01±0.01}                                                    & \multicolumn{1}{c|}{0.16±0.04}                                                                       & 0.17±0.01                                                             \\ \hline
warpPIE10P       & \multicolumn{1}{c|}{\textbf{0.01±0.01}}                                                              & \textbf{0.01±0.01}                                                    & \multicolumn{1}{c|}{\textbf{0.01±0.01}}                                                              & 0.13±0.01                                                             \\ \hline
TOX-171          & \multicolumn{1}{c|}{\textbf{0.02±0.01}}                                                              & \textbf{0.02±0.01}                                                    & \multicolumn{1}{c|}{0.05±0.01}                                                                       & 0.32±0.01                                                             \\ \hline
pixraw10P        & \multicolumn{1}{c|}{\textbf{0.01±0.01}}                                                              & \textbf{0.01±0.01}                                                    & \multicolumn{1}{c|}{0.03±0.01}                                                                       & 0.36±0.01                                                             \\ \hline
CLL-SUB-111      & \multicolumn{1}{c|}{\textbf{0.11±0.03}}                                                              & \textbf{0.17±0.01}                                                    & \multicolumn{1}{c|}{0.12±0.01}                                                                       & 0.39±0.01                                                             \\ \hline
\textbf{Average} & \multicolumn{1}{c|}{\textbf{0.04±0.01}}                                                              & \textbf{0.04±0.01}                                                    & \multicolumn{1}{c|}{0.07±0.02}                                                                       & 0.27±0.01                                                             \\ \hline
\end{tabular}
\label{tab-results-Obj}
\end{center}
\end{table*}

\section{CONCLUSION REMARKS}

The cost of data processing usually increases exponentially when the number of features in a dataset increases. Accordingly, for big data an efficient and fast feature selection method is essential. By considering improving the ratio of retained features as well as the accuracy of classification as our objectives, feature selection becomes a binary-constrained multi-objective optimization problem. We have solved this problem using the novel multi-objective coordinate search method, which generates new individuals by flipping the value of a variable of the solutions in the Pareto front instead of applying crossover and mutation as used in many multi-objective algorithms. In fact, the proposed algorithm investigates the effectiveness of each feature in the task of classification. This method is simple, needs fewer hyper-parameters, uses less memory, and can recognize convergence. Based on the experiment results, our method outperforms NSGA-II, not only in minimizing both objectives in a limited time but also in making the distribution of the Pareto front solutions in the objective space much wider in most cases. The number of features in the resultant optimal sets by the proposed method is significantly less than those obtained by NSGA-II.

As a potential future work, MOCS can be applied to huge-scale datasets, as well as other tasks than feature selection. Also, the non-binary scheme of this method should be developed and tested on a wide range of problems. A unique mark of CS or MOCS is that these algorithms can also be applied for fine-tuning the results of other algorithms, as a local search after their results converge.

\bibliography{ref}
\bibliographystyle{IEEEtran}

\addtolength{\textheight}{-12cm}   

\end{document}